\documentclass[10pt,twocolumn,letterpaper]{article}

\usepackage{iccv}
\usepackage{times}
\usepackage{epsfig}
\usepackage{graphicx}
\usepackage{bm}
\usepackage{amsmath}
\usepackage{amssymb}
\usepackage{amsfonts}
\usepackage{multirow}
\usepackage{float}
\usepackage{caption}
\usepackage{subcaption}
\usepackage{accents}
\usepackage{booktabs} 
\usepackage{enumitem}
\usepackage{color,xcolor,colortbl}
\usepackage[symbol]{footmisc}
\usepackage[accsupp]{axessibility}

\usepackage[pagebackref=true,breaklinks=true,letterpaper=true,colorlinks,bookmarks=false]{hyperref}

\iccvfinalcopy 

\def\iccvPaperID{3141} 

\ificcvfinal\fi

\begin{document}

\title{Single Image Reflection Separation via Component Synergy} 


\author{Qiming Hu \; Xiaojie Guo\thanks{Corresponding author. This work was supported by National Natural Science Foundation of China under Grant no. 62072327.} \\
College of Intelligence and Computing, Tianjin University, Tianjin, China\\
{\tt\small huqiming@tju.edu.cn, xj.max.guo@gmail.com}
}

\maketitle
\ificcvfinal\thispagestyle{empty}\fi


\begin{abstract}
  The reflection superposition phenomenon is complex and widely distributed in the real world, which derives various simplified linear and nonlinear formulations of the problem. In this paper, based on the investigation of the weaknesses of existing models, we propose a more general form of the superposition model by introducing a learnable residue term, which can effectively capture residual information during decomposition, guiding the separated layers to be complete. In order to fully capitalize on its advantages, we further design the network structure elaborately, including a novel dual-stream interaction mechanism and a powerful decomposition network with a semantic pyramid encoder.  Extensive experiments and ablation studies are conducted to verify our superiority over state-of-the-art approaches on multiple real-world benchmark datasets. Our code is publicly available at \hyperlink{https://github.com/mingcv/DSRNet}{https://github.com/mingcv/DSRNet}.
\end{abstract}

\section{Introduction}
\label{sec:introduction}


As a typical layer superimposition scenario, pictures captured through glass-like surfaces may be blended with undesired reflection, which not only impairs the aesthetic value but also hinders the downstream tasks \cite{tog/SinhaKGSS12}. In the meantime, scenes of interest are possibly concealed in reflection \cite{cvpr/WanSLDK20}. Therefore, both the parts transmitted through a surface (transmission layer, $\textbf{T}$) and those reflected (reflection layer, $\textbf{R}$) are desired to be reconstructed from a superimposed image $\textbf{I}$ to fulfill the practical demands.  

\begin{figure}[t]
  \centering
  \includegraphics[width=\linewidth]{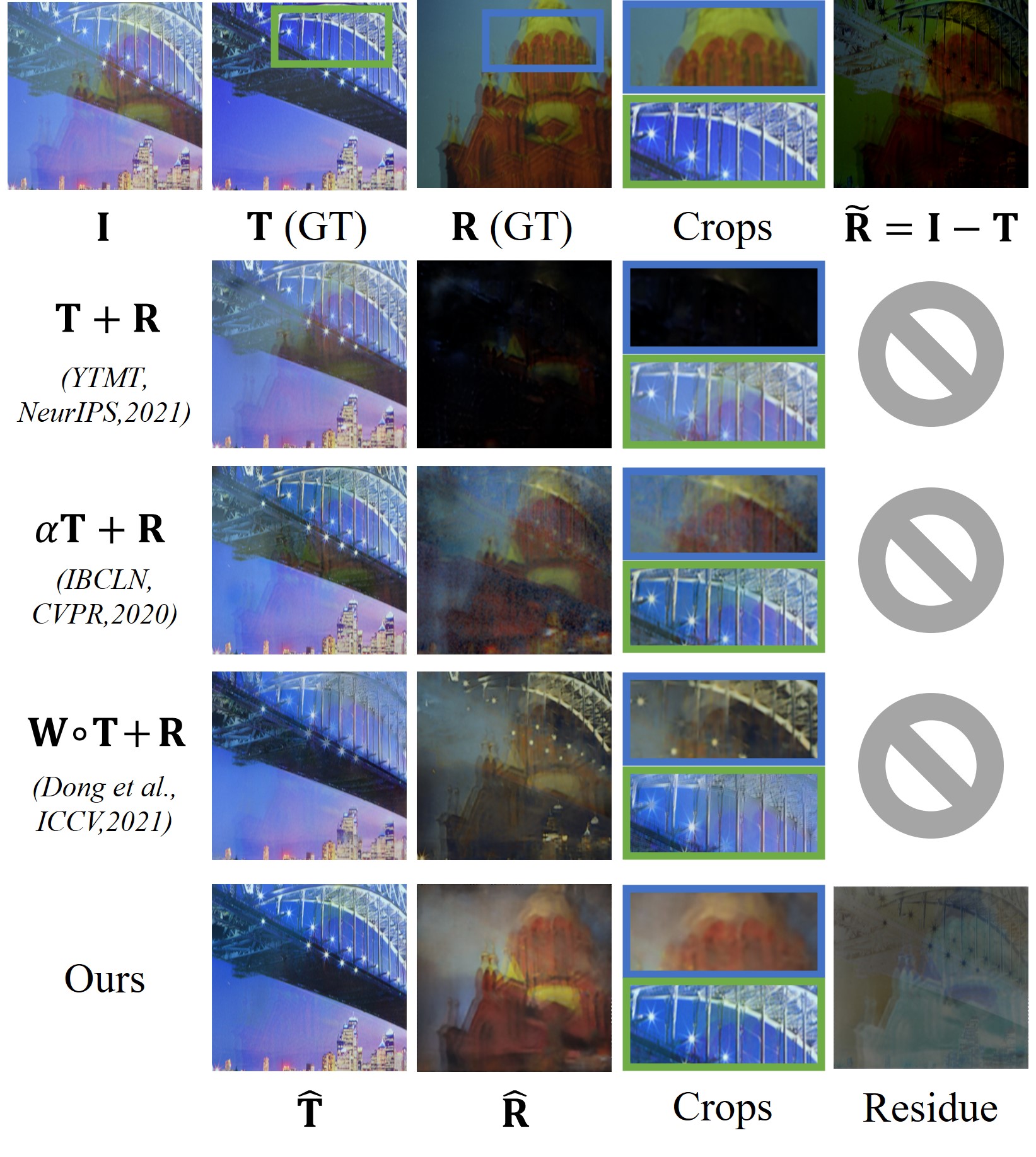}
  \caption{A visual example drawn from the $\text{SIR}^2$ dataset. The first row displays a real input $\textbf{I}$ with ground-truth $\textbf{T}$ and $\textbf{R}$. In rows 2-5, the separations based on different physical models are compared, where $\hat{\textbf{T}}$ and $\hat{\textbf{R}}$ represent the predictions. Our method is able to produce a ``Residue'' term, separating the nonlinearity from the layer reconstructions. }
  \vspace{-10pt}
  \label{fig:residual_term}
\end{figure}

Following a common assumption \cite{pami/LevinW07}, $\textbf{I}$ is linearly composed by $\textbf{T}$ and $\textbf{R}$ expressed as $\textbf{I}=\textbf{T}+\textbf{R}$, which has been popular due to its simplicity. However, in real-world scenarios, the reflection and transmission layers are likely to be weakened due to diffusion and other problems during the superimposition \cite{cvpr/WanSLDK20}. Therefore, several methods \cite{wan2018crrn,eccv/YangGLS18} introduce scalars $\alpha$ and $\beta$ to the two components, respectively, obtaining  $\textbf{I} = \alpha \textbf{T}+\beta \textbf{R}$. However, the linear superimposition model is often violated due to over-exposure and other phenomena \cite{cvpr/WenT0LHH19}. Thus, an alpha-matting map $\textbf{W}$ is introduced to the model as $\textbf{I} = \textbf{W}\circ \textbf{T}+\bar{\textbf{W}}\circ \textbf{R}$ with $\bar{\textbf{W}} = \textbf{1}-\textbf{W}$, which, however, increases the degree of freedom and makes this ill-posed problem much harder. In view of all these limitations, it is not a trivial task to represent different kinds of reflection scenarios with a single model. 
In this work, we take a step forward in advancing the solution and deliver a more general form of the superimposition procedure by introducing a residue term as follows: 
\begin{equation}
    \label{eq:model}
    \textbf{I} = \tilde{\textbf{T}}+\tilde{\textbf{R}} = \textbf{T} + \textbf{R} +\Phi(\textbf{T},\textbf{R}),
\end{equation}
where $\textbf{T}$ and $\textbf{R}$ denote the groundtruth scenes of the transmission and reflection layers, while their respective information contained in $\textbf{I}$ after superimposition and other degradations and finally reach the camera sensors is represented by $\tilde{\textbf{T}}$ and $\tilde{\textbf{R}}$. 
$\Phi(\textbf{T},\textbf{R}) = \textbf{I}-\textbf{T}-\textbf{R}$ is the residue of the reconstruction, which can be caused by attenuation, overexposure, etc. $\Phi(\cdot,\cdot)$ represents a group of functions that can be used to model the residue in specific situations. For example,  $\Phi(\textbf{T},\textbf{R})=0$ in \cite{cvpr/LiB14,iccv/FanYHCW17,pami/LevinW07,cvpr/ZhangNC18a,cvpr/WeiYFW019,nips/HuG21}, $\Phi(\textbf{T},\textbf{R})=(\alpha-1)\textbf{T}+ (\beta-1) \textbf{R}$ in \cite{wan2018crrn,eccv/YangGLS18}, $\Phi(\textbf{T},\textbf{R})=-(\bar{\textbf{W}} \circ \textbf{T}+ \textbf{W} \circ \textbf{R})$ in \cite{cvpr/WenT0LHH19,zheng2020does}. In an effort to rule these different cases, a learnable module is desired to be introduced to model the residue $\Phi(\textbf{T},\textbf{R})$. To clarify it, we provide a visual example in Fig.\;\ref{fig:residual_term}, where the first row shows a superimposed image $\textbf{I}$, its groundtruth transmission $\textbf{T}$ (GT) and reflection $\textbf{R}$ (GT) layers collected by $\text{SIR}^2$ dataset. According to the linear model, the guidance for reflection restoration is  $\tilde{\textbf{R}} = \textbf{I} - \textbf{T}$ with incomplete signals rather than $\textbf{R}$ (GT) that contains full information, which not only impairs the training of networks but also hinders the downstream usages. Therefore, as we point out in Eq.\,(\ref{eq:model}), a promising way out of this dilemma is to utilize ground-truth layers $\textbf{T}$ and $\textbf{R}$ as guiding signals and employ an extra residual term to handle the non-linearity (``Residue'' in Fig.\,\ref{fig:residual_term}). As shown, the components that violate the linear assumption are separated from the layer predictions $\hat{\textbf{T}}$ and $\hat{\textbf{R}}$, and thus more complete and precious layer separations than previous methods are obtained.

To further exploit the synergy between the components $\textbf{T}$ and $\textbf{R}$, we deliberate upon the facilitation of inter-component feature interaction. The efficiency of a dual-stream interactive network has been verified in SIRS problem by \cite{nips/HuG21}. Building upon their analysis, we subsequently advance a more effective scheme. Inspired by the gated mechanisms \cite{eccv/ChenCZS22,tu2022maxim}, we present a simple yet effective mutually-gated interaction diagram (MuGI) serving as a better feature interaction candidate, by means of which we design the MuGI block to build our decomposition network.
Moreover, the hypercolumn \cite{cvpr/ZhangNC18a}, as a commonly used semantic information encoder in the SIRS task, tends to aggregate features in a lossy way. Therefore, we propose a dual-stream pyramid fusion network (DSFNet) to replace it, which hierarchically decomposes and fuses the multi-scale semantic information leveraging our proposed MuGI blocks and dual-stream fusion blocks (DSF Block). The roughly decomposed features of layers are further delivered into the dual-stream fine-grained decomposition phase (DSDNet). The two sub-networks constitute the main branch of our \textbf{D}ual-stream \textbf{S}emantic-aware network with \textbf{R}esidual correction, namely DSRNet.

In summary, our main contributions are as follows:
\begin{itemize}[itemsep=0pt]
    \item We build a general form for SIRS by introducing a learnable residue term, which is more flexible to different scenarios and boosts the separation of layers;
    \item We exploit the synergy of features via mutually-gated interaction block within a dual-stream semantic-aware network, which facilitates information usage;
    \item Extensive experiments are conducted to demonstrate the effectiveness of our design. Overall, it achieves state-of-the-art performance against the alternatives on multiple real-world benchmarks for SIRS.
\end{itemize}

 \section{Related Work}
Plenty of efforts have been devoted to handling the image reflection separation problem in the past decades. They either leverage multiple images to acquire more cues for the layer decomposition or exploit extra priors to manage the problem through a single image.  In what follows, we organize the previous methods based on the images they require.

\textbf{Multiple Image Reflection Separation}. In general, the transmission component tends to be unpolarized, yet the reflection component varies when rotating the polarization filter mounted in front of a camera sensor. Inspired by this phenomenon, a variety of methods \cite{ijcv/NayarFB97,tip/KongTS11,pami/KongTS14,nips/LyuCLPS19,eccv/LiQZH20,cvpr/LeiHZYSC20} resort to the physical solution that separates different components through a sequence of images with varied polarization orientations. Besides the polarization cues, the layer separations can also be indicated by different focuses \cite{cvpr/FaridA99}, stereo information \cite{tog/SinhaKGSS12}, flash on/off \cite{tog/AgrawalRNL05,cvpr/LeiC21}, relative motions \cite{iccv/SarelI05,pami/GaiSZ12,iccv/LiB13,tog/Freeman15,cvpr/YangLDT16,cvpr/LiuL0CH20,pami/LiuLYCH22} and scene consistency \cite{cvpr/GuoCM14,cvpr/SimonP15,cvpr/HanS17}.
Although involving multiple images mitigates the ill-posedness of the separation problem, \emph{this pipeline requires professional devices (such as polarizers) and manual operations, which limits its application. }

\begin{figure*}[t]
	\centering
	\includegraphics[width=\linewidth]{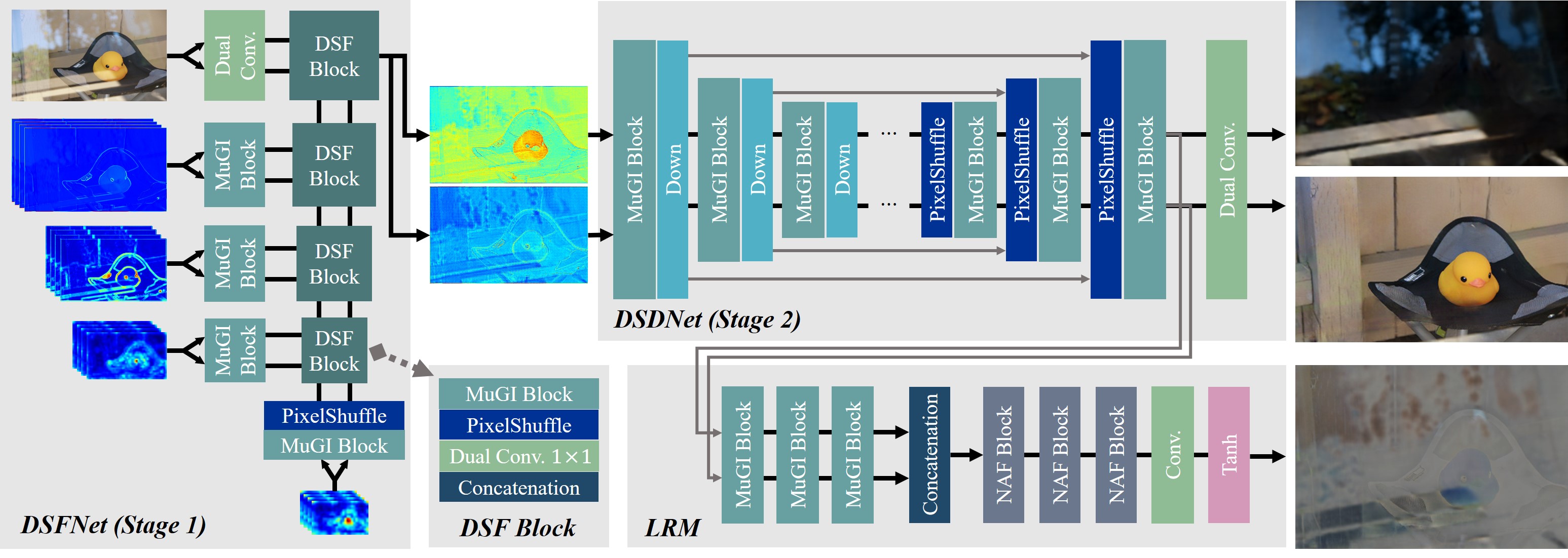}
	\caption{The architecture of our proposed DSRNet, which consists of two cascaded stages and a learnable residue module (LRM). In Stage 1, DSFNet aggregates and separates hierarchical semantic information with two interactive feature streams. The roughly separated features are further fed into the DSDNet for fine-grained decompositions in Stage 2, where the LRM takes decomposed features, separating out the components that violate the linear assumption. }
	\label{fig:arch}
\end{figure*}

\textbf{Single Image Reflection Separation}. Compared with multiple-image solutions, single-image-based methods show more merits, being more flexible and requiring less manual operations.
However, as there is no such thing as a free lunch, single-image schemes call for additional priors to cope with its ill-posedness nature, such as gradient-based constraints \cite{cvpr/LevinZW04,nips/LevinZW02,pami/LevinW07,cvpr/LiB14} and  manual annotations \cite{pami/LevinW07}.  Specifically, Levin \emph{et al.} \cite{cvpr/LevinZW04, nips/LevinZW02} impose sparse constraints on gradients to acquire the separations that have fewer edges. Levin and Weiss \cite{pami/LevinW07} demand several manual annotations of respective layers to obtain favorable decompositions, which involves additional human labor. Li and Brown \cite{cvpr/LiB14} suppose that one layer is smoother than the other, thus applying unbalanced penalization on the gradients of the two layers to decompose them. \emph{Although these methods work fine in restricted scenarios, the complicated real-world conditions can considerably surpass their assumptions, leading to unsatisfactory results. } 

 
 
The advent of deep learning technology mitigates the incongruity between the volume of data and modeling capacity. Certain underlying assumptions can be subtly embedded in the process of data synthesis, and subsequently learned by deep models. Concretely, CEILNet \cite{iccv/FanYHCW17} imposes the relative smoothness prior to the synthesis of reflection layers and combines them with transmission layers by addition. An edge-aware network is developed to capture the transmission components, which yet ignores high-level semantics that are likely to aid the SIRS task. Zhang \emph{et al.} \cite{cvpr/ZhangNC18a} hence introduce HyperColumn features \cite{cvpr/HariharanAGM15} using a pre-trained VGG-19 network to acquire semantic awareness, besides the perceptual and adversarial losses. Moreover, the exclusivity loss is developed to penalize the intersected gradients.  ERRNet \cite{cvpr/WeiYFW019} goes a step further by leveraging an additional set of misaligned pairs. However, it appears to overlook the estimation of the reflection layer, which is essentially an image component and potentially important to distinguish the transmission parts. Li \emph{et al.} \cite{corr/abs-2012-00945} thereby presents a two-stage network (RAGNet) to first estimate reflection components and predict transmission ones guided by them, whereas the reflection estimation is isolated from the transmission in this way. Further, Hu and Guo \cite{nips/HuG21} come up with the YTMT strategy, which pays equal attention to the two components, and a dual-stream interactive network is developed to restore the two layers simultaneously.  However, their linear assumption makes the predicted reflection components tend to be weak in many cases.
Other than them, BDN \cite{eccv/YangGLS18} and IBCLN \cite{cvpr/LiY0LH20} make use of reflection models weighted by scalars, and iteratively estimate both the components. This scheme prevents the reflection from being too weak, but its restorations often struggle to be free from the transmission parts.  Furthermore, Wen \emph{et al.} \cite{cvpr/WenT0LHH19} simulate the nonlinear superimposition phenomenon by predicting a three-channel alpha blending weight map with the adversarial guidance of collected unpaired images. Dong \emph{et al.} \cite{iccv/Dong00BXL21} develops an iterative network and estimates a probabilistic reflection confidence map in each iteration. \emph{However, it is not an easier task to estimate an alpha blending map than to determine each of the blended components, while the residual estimation only needs to retain the redundant information during decomposition, which is more practical than the former.} Considering the drawbacks of the previous reflection models, we propose a general form of them with a learnable residue term, which is shown to be more effective and flexible.


\section{Methodology}
As shown in Fig.\;\ref{fig:arch}, our proposed DSRNet comprises two cascaded stages and a learnable residue module (LRM). In what follows, we first clarify the motivation of LRM and then introduce the MuGI block and the architecture of the DSRNet followed by its training loss functions.


\subsection{Residue of the Linear Combination}
\label{sec:residue_term}
It often occurs that a superimposed image $\textbf{I}$ cannot be perfectly represented by the linear combination of $\textbf{T}$ and $\textbf{R}$, leading to a residue term, as discussed in Section \ref{sec:introduction}. Under its disturbance, the error of the additive reconstruction criterion $\varepsilon = \|\textbf{I}-(\hat{\textbf{T}}+\hat{\textbf{R}})\|$ remains large even though the predicted layers can well reconstruct the ground truths. Further minimizing $\varepsilon$ will put redundant information to either $\hat{\textbf{T}}$ or $\hat{\textbf{R}}$, which induces the deviation of predictions from their ground truths instead. Moreover, there are many kinds of simplified physical models proposed by previous work with their own shortcomings. Naturally, a unified model is desired to be put forward. Therefore, we introduce an extra residue term to offset the error in additive reconstruction, which, at the same time, unifies the different physical models. As depicted in Fig.\;\ref{fig:arch}, we leverage a learnable residue module (LRM) after Stage 2 to estimate the residual information during the decomposition by analyzing the features of $\hat{\textbf{T}}$ and $\hat{\textbf{R}}$ before the final layer. LRM consists of an interactive and a fusion part to collect and merge residual information from two branches. The dual-stream signals are concatenated before entering the fusion part. To constrain the space of residue, the $\text{tanh}(\cdot)$ function is utilized as the final activation. With the participation of LRM, we define the following reconstruction loss with residual rectification ($\text{R}^3$ Loss):
\begin{equation}
    \mathcal{L}_{rec} := \|\textbf{I} - (\hat{\textbf{T}}+\hat{\textbf{R}}) - \Phi(\hat{\textbf{T}},\hat{\textbf{R}})\|_1,
\end{equation}
where $\Phi$ denotes the LRM and $\|\cdot\|_1$ means the $\ell_1$ norm. Guided by this objective, the information beyond the additive model will flow to the residue term. Notably, this term can be totally discarded during the testing phase, avoiding extra computational costs. 

\subsection{Mutually-gated Interactive Block}

\begin{figure}[t]
	\centering
	\includegraphics[width=\linewidth]{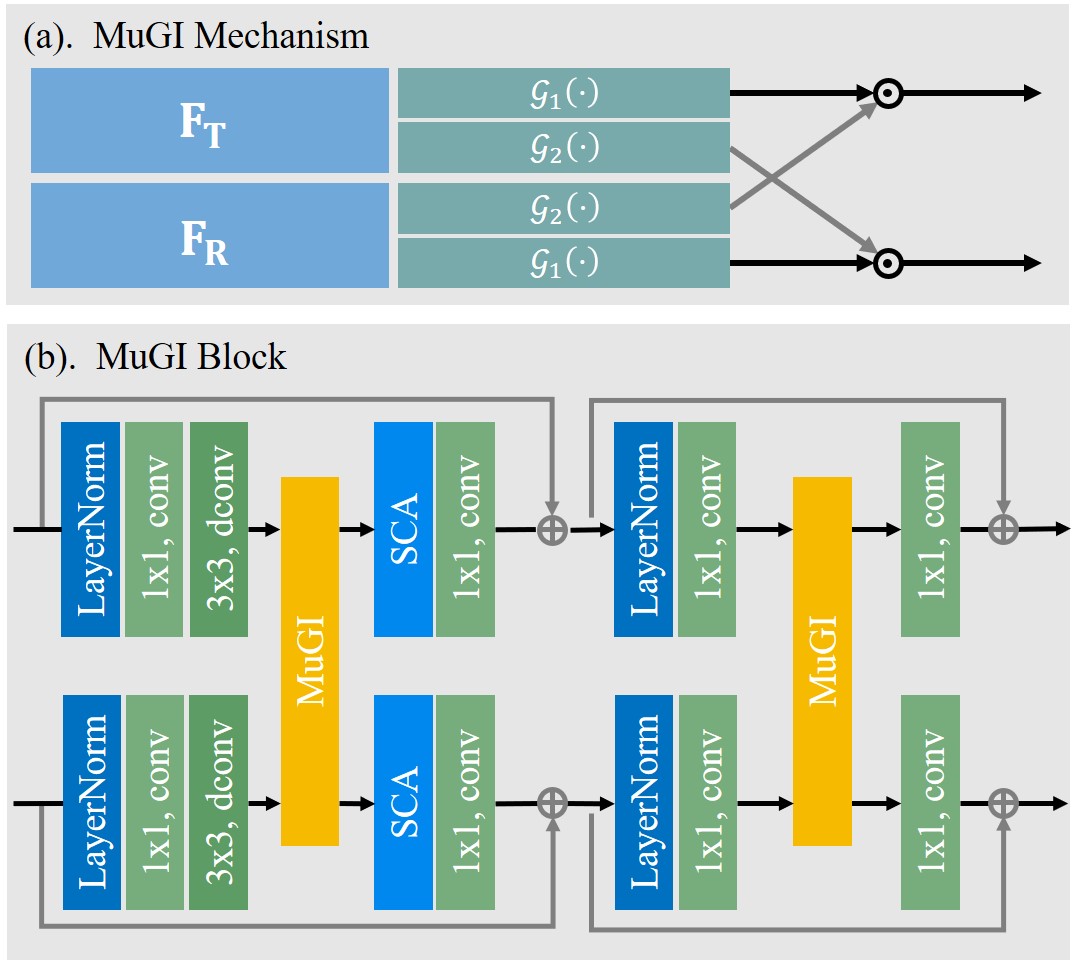}
	\caption{MuGI mechanism and MuGI Block.}
	\label{fig:block}
\end{figure}

Now that we employ the LRM to model the residual components during the additive reconstruction, the rest part can be seen as a linear model again as $\tilde{\textbf{I}} = \textbf{I} - \Phi(\hat{\textbf{T}}, \hat{\textbf{R}}) = \hat{\textbf{T}} + \hat{\textbf{R}}$. As displayed in Fig.\;\ref{fig:block}\;(a), we exploit the synergy between $\hat{\textbf{T}}$ and $\hat{\textbf{R}}$ by developing the \textbf{Mu}tually-\textbf{G}ated \textbf{I}nteractive mechanism, namely MuGI:
\begin{equation}
\left\{
\begin{aligned}
\hat{\textbf{F}}_{\textbf{T}} &= \mathcal{G}_1(\textbf{F}_\textbf{T}) \circ \mathcal{G}_2(\textbf{F}_\textbf{R}); \\
\hat{\textbf{F}}_{\textbf{R}} &= \mathcal{G}_1(\textbf{F}_\textbf{R}) \circ \mathcal{G}_2(\textbf{F}_\textbf{T}),
\end{aligned}
\right.
\label{eq:mutual_gate}
\end{equation}
where $\textbf{F}_\textbf{T}, \textbf{F}_\textbf{R} \in \mathbb{R}^{H_1 \times W_1 \times C_1}$ denote intermediate features of streams for predicting $\hat{\textbf{T}}$ and $\hat{\textbf{R}}$, respectively, and $\hat{\textbf{F}}_\textbf{T}, \hat{\textbf{F}}_\textbf{R} \in \mathbb{R}^{H_2 \times W_2 \times C_2}$ represent the outputs of the mutual gate. $\mathcal{G}_1$ and $\mathcal{G}_2$ are functions selecting which part of features to engage the interaction. $\circ$ indicates the element-wise multiplication. Here we use a simple implementation of MuGI to evaluate its capability of solving SIRS, which divides a feature map in half at the channel dimension, and $\mathcal{G}_1(\cdot)$ always selects the former half, while $\mathcal{G}_2(\cdot)$ chooses the latter. We have $H_1=H_2,W_1=W_2$ and $C_2=C_1/2$ in this way.   Such a gated mechanism captures the mutual dependency of the two components in the feature space, say if either stream contains signals the other one desires, they tend to be split into the latter half (selected by $\mathcal{G}_2(\cdot)$) and transferred into the sibling stream by the gate.

Our simple implementation of MuGI makes it plug-and-play.  Therefore, in general, most of the single-path  blocks can be converted into a dual-stream one by parallel setting two of them and relating them with MuGI. In this paper, we follow a state-of-the-art block design presented by \cite{eccv/ChenCZS22} and convert its single-path NAF block into a dual-stream mutually-gated interactive block (MuGI block). Its diagram is depicted in Fig.\;\ref{fig:block}\;(b), in which the simple gates are replaced by MuGI gates. Specifically, the layer normalization makes dual-stream features comparable, $1\times1$ convolution doubles the channel dimensions in order to conduct MuGI without information loss which reduces dimensions by half during the interaction, channel attention and the following $1\times1$ convolution serve as fusion/reweighting roles. 


\begin{table*}[t]
  \centering
  \begin{tabular}{ccccccccccc}
  \toprule[1pt]
  \multirow{2}{*}{Methods} & \multicolumn{2}{c}{Real20 (20)} & \multicolumn{2}{c}{Objects (200)} & \multicolumn{2}{c}{Postcard (199)} &\multicolumn{2}{c}{Wild (55)} & \multicolumn{2}{c}{Average} \\ \cline{2-11}
  & PSNR & SSIM & PSNR  & SSIM & PSNR & SSIM & PSNR & SSIM & PSNR & SSIM \\ \hline
  Zhang \emph{et al.} \cite{cvpr/ZhangNC18a} & 22.55 & 0.788 & 22.68 & 0.879 & 16.81 & 0.797 & 21.52 & 0.832 & 20.08 & 0.835 \\
  BDN \cite{eccv/YangGLS18}  & 18.41 & 0.726 & 22.72 & 0.856 & 20.71 & 0.859 & 22.36 & 0.830 & 21.65 & 0.849 \\
  ERRNet \cite{cvpr/WeiYFW019} & 22.89 & 0.803 & 24.87 & 0.896 & 22.04 & 0.876 & 24.25 & 0.853 & 23.53 & 0.879 \\
  IBCLN \cite{cvpr/LiY0LH20} & 21.86 & 0.762 & 24.87 & 0.893 & 23.39 & 0.875 & 24.71 & 0.886 & 24.10 & 0.879 \\
  RAGNet \cite{corr/abs-2012-00945} & 22.95 & 0.793 & 26.15 & 0.903 & 23.67 & 0.879 & 25.53 & 0.880 & 24.90 & 0.886 \\
  DMGN \cite{tip/FengPJCZL21} & 20.71 & 0.770 & 24.98 & 0.899 & 22.92 & 0.877 & 23.81 & 0.835 & 23.80 & 0.877 \\
  Zheng \emph{et al.} \cite{cvpr/ZhengSCJDK21} & 20.17 & 0.755 & 25.20 & 0.880 & 23.26 & 0.905 & 25.39 & 0.878 & 24.19 & 0.885 \\
  YTMT \cite{nips/HuG21} & 23.26 & 0.806 & 24.87 & 0.896 & 22.91 & 0.884 & 25.48 & 0.890 & 24.05 & 0.886 \\
  Ours & \textbf{24.23} & \textbf{0.820} & \textbf{26.28} & \textbf{0.914} & \textbf{24.56} & \textbf{0.908} & \textbf{25.68} & \textbf{0.896} & \textbf{25.40} & \textbf{0.905} \\ \hline
  ${\text{Dong \emph{et al.}}}^\dagger$ \cite{iccv/Dong00BXL21} & 23.34 & 0.812 & 24.36 & 0.898 & 23.72 & 0.903 & 25.73 & 0.902 & 24.21 & 0.897 \\

  $\text{Ours}^\dagger$ & \textbf{23.91} & \textbf{0.818} & \textbf{26.74} & \textbf{0.920} & \textbf{24.83} & \textbf{0.911} & \textbf{26.11} & \textbf{0.906} & \textbf{25.75} & \textbf{0.910} \\
  \bottomrule[1pt]
  \end{tabular}
  \caption{Quantitative results on four real-world benchmark datasets of methods. The best results are indicated in \textbf{bold}. $\dagger$ indicates extra training data that are involved as in \cite{iccv/Dong00BXL21}. }
  \label{tab:qcomp}
\end{table*}

\subsection{Dual-stream Semantic-aware Network}


The semantic information provided by pre-trained models is usually introduced to alleviate the impact of ill-posedness in SIRS. The HyperColumn \cite{cvpr/ZhangNC18a,cvpr/WeiYFW019,nips/HuG21}, as a prevalent multi-scale semantic feature extractor in SIRS, aggregates semantics through interpolating the features extracted by a pre-trained model into the same scale as the input images. It then concatenates them together and employs a $1 \times 1$ convolution to rapidly reduce the channel dimensions (typically from 1475 to 64 or 256), before feeding them into the decomposition networks. This strategy omits the inner relationship of multi-scale features and may discard informative signals. To overcome the drawbacks mentioned above, we present a dual-stream pyramid fusion network (DSFNet), which hierarchically aggregates the extracted features by jointly upsampling and interacting. 
As shown in Fig.\;\ref{fig:arch}, given the hierarchical features extracted by a pre-trained deep network (e.g., VGGNet-19), DSFNet gathers them in a bottom-up manner with MuGI blocks and dual-stream fusion blocks (DSF Block), the structure of which is shown in Fig.\;\ref{fig:arch}. ``Dual Conv'' means two parallel convolutional layers. The features extracted by the pre-trained network at each level first interact through a MuGI block, thereby features in the same scale are related and transformed into dual-stream features ($\textbf{F}_\textbf{T}=\textbf{F}_\textbf{R}$ for these blocks). After the interaction, the deepest features extracted by VGGNet are upscaled and then fused with shallower features through the DSF block, in which the dual-stream features come from the two adjacent scales are concatenated at channel dimension, fused by a 1x1 convolutional layer and then upscaled at each stream. The MuGI blocks are further employed to promote cross-scale interactions. Note that, shallow features extracted from RGB inputs are fused at the top of DSFNet, which preserves fine-grained details to restore both layers. The aggregated features provide rough separations of layers and are further refined by the second stage. The following stage (DSDNet) is constructed by the proposed MuGI blocks in a U-shaped manner. More details can be found in the supplementary materials. 



\subsection{Loss Function}

Besides the $\text{R}^3$ loss demonstrated in Section \ref{sec:residue_term}, we further introduce pixel, perceptual, and exclusion loss for pixel-wise and semantic fidelity as well as the gradient independence of layers, which are described as follows: 

\noindent\textbf{Pixel Loss.} The pixel loss is used to constrain the consistency of the reconstructed layer and groundtruth in both the natural image domain and gradient domain, minimizing their errors as follows:
\begin{equation}
\begin{aligned}
     \mathcal{L}_{pix} := & \,\| \hat{\textbf{T}} - \textbf{T} \|^2_2 + \|\hat{\textbf{R}} - \textbf{R}\|^2_2 \\
    + & \,\alpha(\|\nabla \hat{\textbf{T}}-\nabla \textbf{T}\|_{1}+\|\nabla \hat{\textbf{R}}-\nabla \textbf{R}\|_{1}),
\end{aligned}
\end{equation}
where $\|\cdot\|_2$ denotes the $\ell_2$ norm. $\nabla$ stands for the gradient operator, which extracts gradients in both vertical and horizontal directions. We express one of them in the formulation for clarity and average them in practice.  $\alpha$ is set as $2$ in our experiments. 

\noindent\textbf{Perceptual Loss.} The pixel-wise losses cannot assure the multi-scale consistency between predictions and their groundtruths, which may result in overwhelming punishment for the whole image due to small differences in local brightness. Therefore, the perceptual loss is further introduced as follows:
\begin{equation}
\mathcal{L}_{per} := \sum_{i} \omega_{i}\|\phi_{i}(\hat{\textbf{T}})-\phi_{i}(\textbf{T})\|_{1},
\end{equation}
where $\phi_i(\cdot)$ designates the features drawn by layer $i\in\{2, 7, 12, 21, 30\}$ of a VGG-19 model. $\omega_i$s are combining weights of terms at different layers. We follow the setting of hyper-parameters in \cite{cvpr/WeiYFW019}.


\noindent\textbf{Exclusion Loss.} We introduce the exclusion loss to strengthen the gradient independence prior and reduce the structural coupling in estimated separations as below:
\begin{equation}
\begin{aligned}
    \mathcal{L}_{exc} &:= \frac{1}{N} \sum_{n=0}^{N-1} \|\Psi(\hat{\textbf{T}}^{\downarrow n}, \hat{\textbf{R}}^{\downarrow n})\|_2^2, \\
    \Psi(\hat{\textbf{T}}, \hat{\textbf{R}})&:=\tanh \left(\eta_{1}|\nabla \hat{\textbf{T}}|\right) \circ \tanh \left(\eta_{2}|\nabla \hat{\textbf{R}}|\right),
\end{aligned}
\end{equation}
where $\hat{\textbf{T}}^{\downarrow n}$ and $\hat{\textbf{R}}^{\downarrow n}$ represent taking down-sampling by $2^n$ times ($2^N$ at most) of $\hat{\textbf{T}}$ and $\hat{\textbf{R}}$. $\eta_1$ and $\eta_2$ are normalization factors, which are identical to \cite{cvpr/ZhangNC18a}. 

Gathering all the loss terms yields the final objective as:
\begin{equation}
    \mathcal{L}_{all} := \mathcal{L}_{pix} + \beta_1\mathcal{L}_{per} + \beta_2\mathcal{L}_{exc} + \beta_3\mathcal{L}_{rec},
\end{equation}
where $\beta_1 = 0.01$, $\beta_2 = 1$, and $\beta_3 = 0.2$ are set empirically.

\begin{figure*}[t]
     \centering{
     \begin{subfigure}{0.18\linewidth}
          \includegraphics[width=1\linewidth,height=60pt]{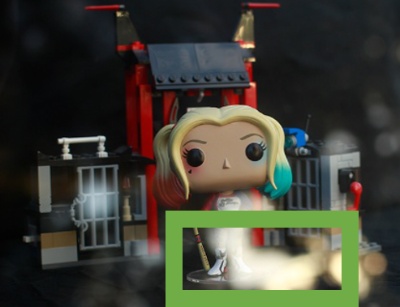}
          \includegraphics[width=1\linewidth,height=37.5pt]{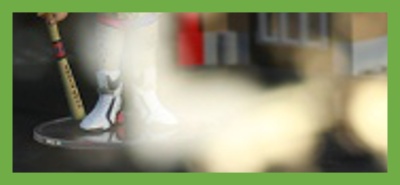}\vspace{3pt}
          \includegraphics[width=1\linewidth,height=60pt]{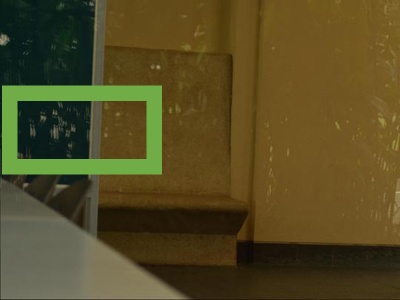}
          \includegraphics[width=1\linewidth,height=37.5pt]{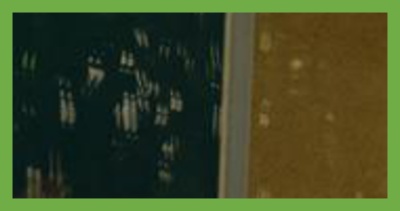}
          \subcaption*{Input}
     \end{subfigure}
     \begin{subfigure}{0.18\linewidth}
          \includegraphics[width=1\linewidth,height=60pt]{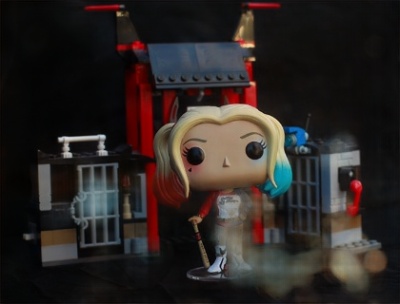}
          \includegraphics[width=1\linewidth,height=37.5pt]{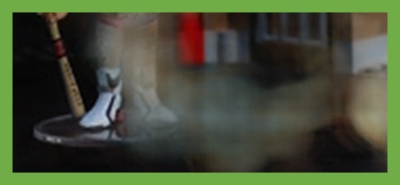}\vspace{3pt}
          \includegraphics[width=1\linewidth,height=60pt]{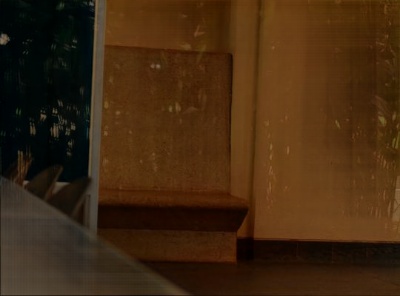}
          \includegraphics[width=1\linewidth,height=37.5pt]{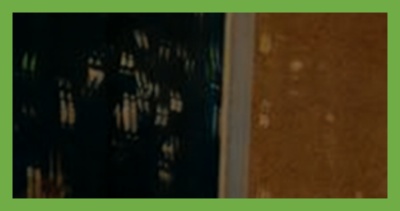}
          \subcaption*{Zhang \emph{et al.} \cite{cvpr/ZhangNC18a}}
    \end{subfigure}
     \begin{subfigure}{0.18\linewidth}
          \includegraphics[width=1\linewidth,height=60pt]{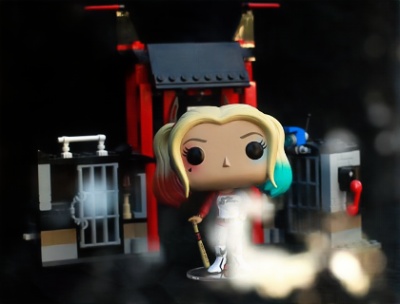}
          \includegraphics[width=1\linewidth,height=37.5pt]{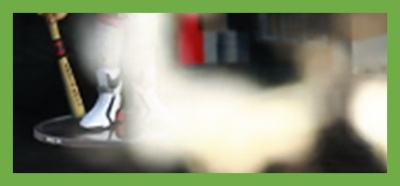}\vspace{3pt}
          \includegraphics[width=1\linewidth,height=60pt]{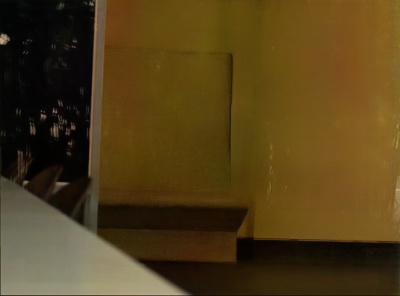}
          \includegraphics[width=1\linewidth,height=37.5pt]{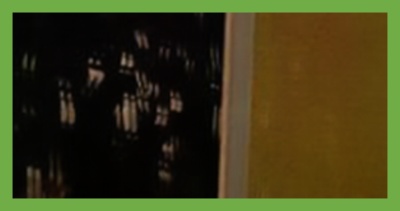}
          \subcaption*{BDN \cite{eccv/YangGLS18}}
    \end{subfigure}
     \begin{subfigure}{0.18\linewidth}
          \includegraphics[width=1\linewidth,height=60pt]{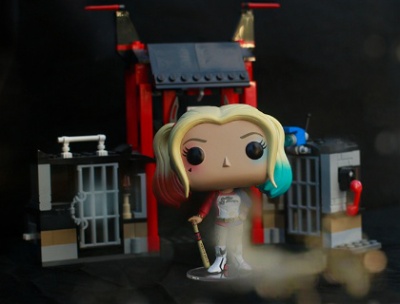}
          \includegraphics[width=1\linewidth,height=37.5pt]{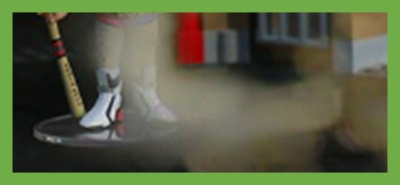}\vspace{3pt}
          \includegraphics[width=1\linewidth,height=60pt]{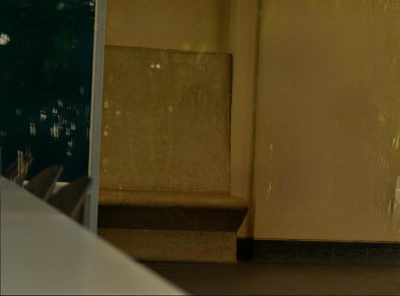}
          \includegraphics[width=1\linewidth,height=37.5pt]{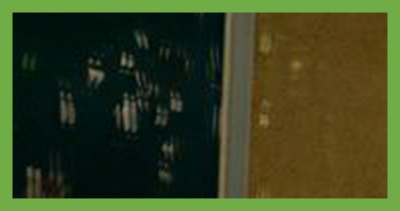}
          \subcaption*{ERRNet \cite{cvpr/WeiYFW019}}
    \end{subfigure}
     \begin{subfigure}{0.18\linewidth}
          \includegraphics[width=1\linewidth,height=60pt]{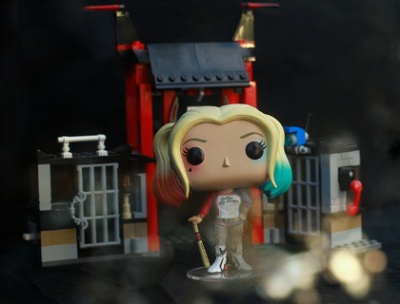}
          \includegraphics[width=1\linewidth,height=37.5pt]{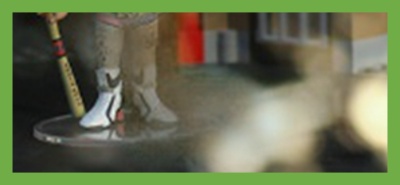}\vspace{3pt}
          \includegraphics[width=1\linewidth,height=60pt]{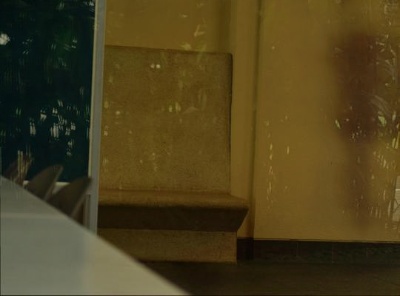}
          \includegraphics[width=1\linewidth,height=37.5pt]{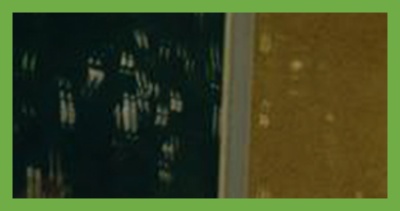}
          \subcaption*{IBCLN \cite{cvpr/LiY0LH20}}
    \end{subfigure}
     \begin{subfigure}{0.18\linewidth}
          \includegraphics[width=1\linewidth,height=60pt]{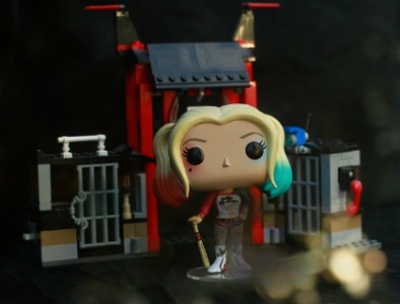}
          \includegraphics[width=1\linewidth,height=37.5pt]{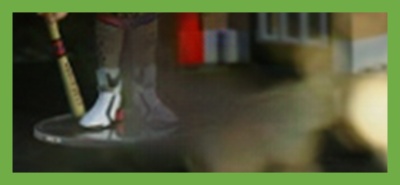}\vspace{3pt}
          \includegraphics[width=1\linewidth,height=60pt]{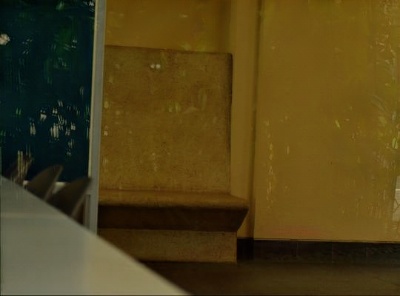}
          \includegraphics[width=1\linewidth,height=37.5pt]{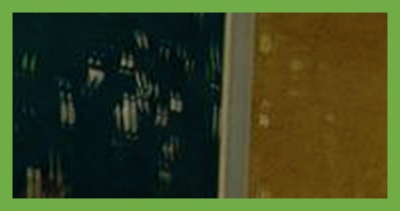}
          \subcaption*{RAGNet \cite{corr/abs-2012-00945}}
    \end{subfigure}
     \begin{subfigure}{0.18\linewidth}
          \includegraphics[width=1\linewidth,height=60pt]{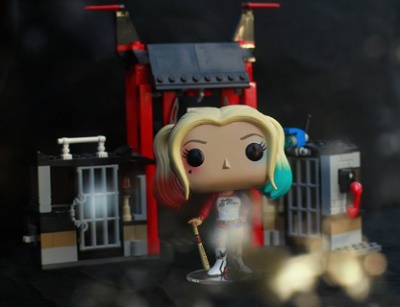}
          \includegraphics[width=1\linewidth,height=37.5pt]{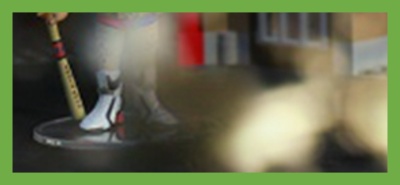}\vspace{3pt}
          \includegraphics[width=1\linewidth,height=60pt]{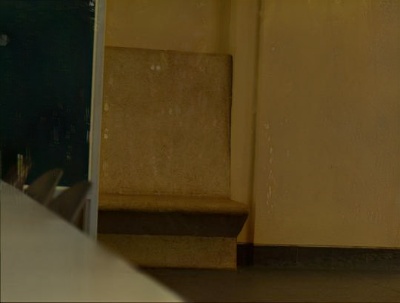}
          \includegraphics[width=1\linewidth,height=37.5pt]{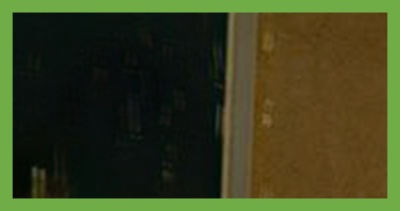}
          \subcaption*{YTMT \cite{nips/HuG21}}
    \end{subfigure}
    \begin{subfigure}{0.18\linewidth}
          \includegraphics[width=1\linewidth,height=60pt]{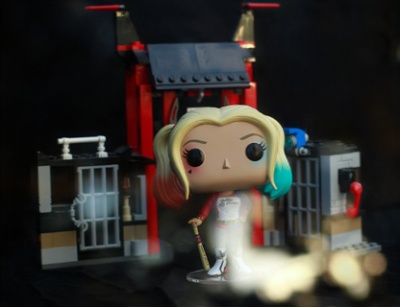}
          \includegraphics[width=1\linewidth,height=37.5pt]{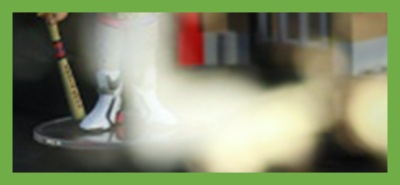}\vspace{3pt}
          \includegraphics[width=1\linewidth,height=60pt]{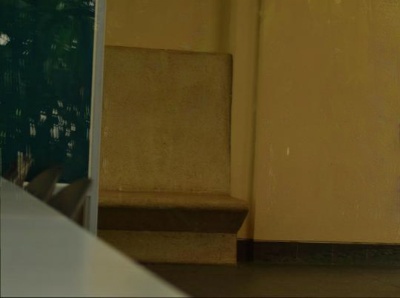}
          \includegraphics[width=1\linewidth,height=37.5pt]{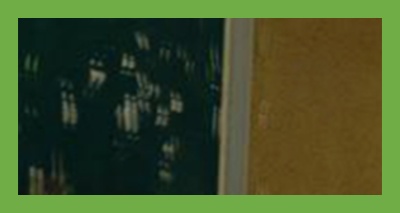}
          \subcaption*{Dong \emph{et al.} \cite{iccv/Dong00BXL21}}
    \end{subfigure}
     \begin{subfigure}{0.18\linewidth}
          \includegraphics[width=1\linewidth,height=60pt]{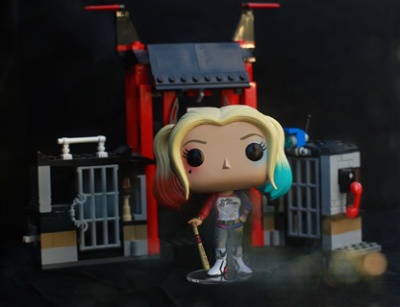}
          \includegraphics[width=1\linewidth,height=37.5pt]{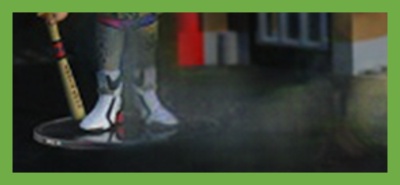}\vspace{3pt}
          \includegraphics[width=1\linewidth,height=60pt]{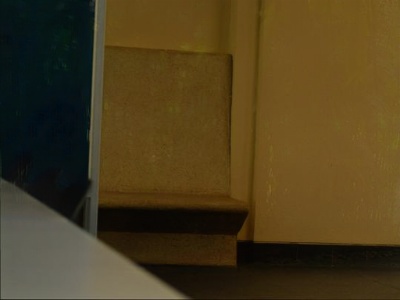}
          \includegraphics[width=1\linewidth,height=37.5pt]{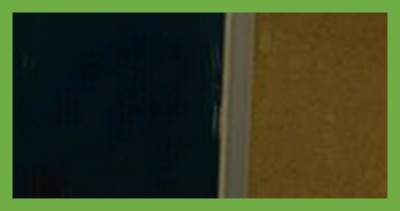}
          \subcaption*{Ours}
    \end{subfigure}
     \begin{subfigure}{0.18\linewidth}
          \includegraphics[width=1\linewidth,height=60pt]{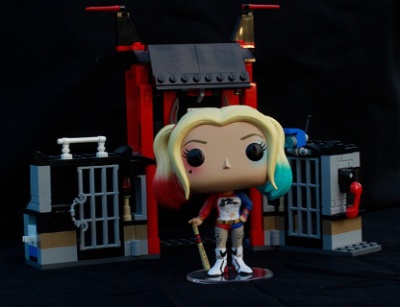}
          \includegraphics[width=1\linewidth,height=37.5pt]{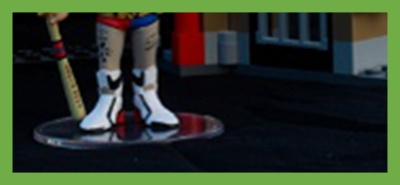}\vspace{3pt}
          \includegraphics[width=1\linewidth,height=60pt]{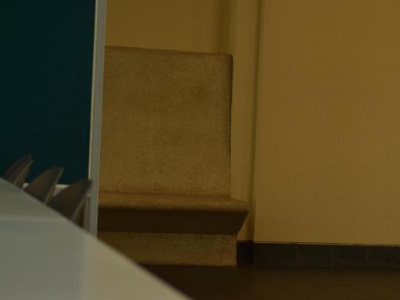}
          \includegraphics[width=1\linewidth,height=37.5pt]{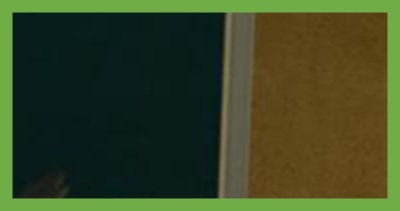}
          \subcaption*{GT}
    \end{subfigure}}
     \caption{Visual comparison of estimated transmission layers between state-of-the-arts and ours on real-world samples.} 
     \label{fig:visual_comp}
\end{figure*}

\section{Experimental Validation}
\subsection{Implementation Details}
Implemented in PyTorch, our models are optimized with the Adam optimizer on a single RTX 3090 GPU for 20 epochs at most to reach favorable overall performance.
The learning rate is initialized as $10^{-4}$ and fixed during the training phase with a batch size of 1. 


\noindent\textbf{Dataset.} Our training dataset embraces both real and synthesized images. Following \cite{iccv/FanYHCW17,nips/HuG21}, the training dataset are composed of 90 real pairs from \cite{cvpr/ZhangNC18a} and 7,643 synthesized pairs from the PASCAL VOC dataset \cite{ijcv/EveringhamGWWZ10}. For synthesized data, transmission and reflection layers are weakened by coefficients $\gamma_1\in[0.8,1.0]$ and $\gamma_2\in[0.4,1.0]$ during blending them with the following model: 
\begin{equation}
    \textbf{I}_{syn} = \gamma_1 \textbf{T}_{syn} + \gamma_2 \textbf{R}_{syn} - \gamma_1\gamma_2 \textbf{T}_{syn} \circ \textbf{R}_{syn},
\end{equation}
where $\textbf{T}_{syn}, \textbf{R}_{syn}$ and $\textbf{I}_{syn}$ represent the transmission, reflection, and superimposed layers during synthesis, respectively. This formulation is inspired by the ``screen'' blending mode in digital image processing, which always reserves lighter colors for the blending layers.

\begin{figure*}[t]
     \centering{
     \hspace{-5pt}
     \begin{subfigure}{0.185\linewidth}
          \includegraphics[width=1\linewidth,height=60pt]{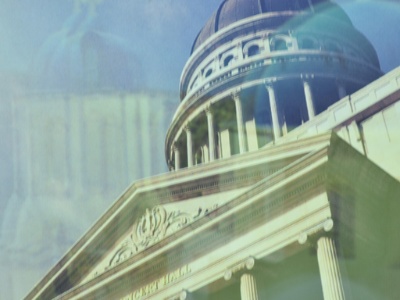}
          \subcaption*{Input}
     \end{subfigure}
     \begin{subfigure}{0.185\linewidth}
          \includegraphics[width=1\linewidth,height=60pt]{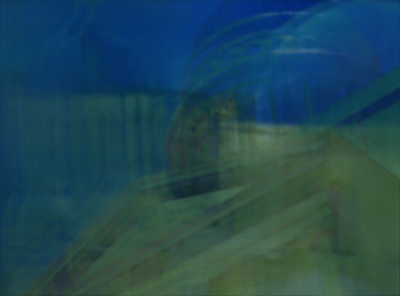}
          \subcaption*{Zhang \emph{et al.} \cite{cvpr/ZhangNC18a}}
     \end{subfigure}
     \begin{subfigure}{0.185\linewidth}
          \includegraphics[width=1\linewidth,height=60pt]{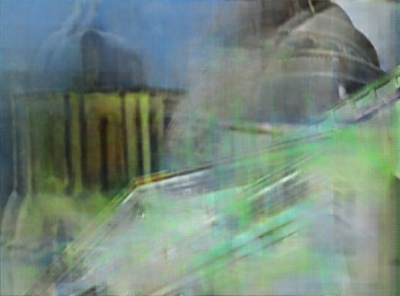}
          \subcaption*{BDN \cite{eccv/YangGLS18}}
     \end{subfigure}
     \begin{subfigure}{0.185\linewidth}
          \includegraphics[width=1\linewidth,height=60pt]{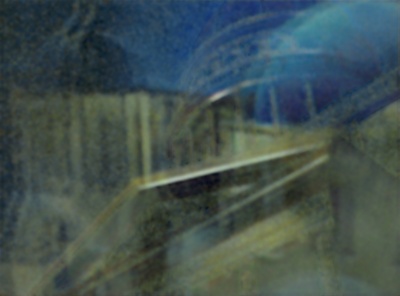}
          \subcaption*{IBCLN \cite{cvpr/LiY0LH20}}
     \end{subfigure}
     \begin{subfigure}{0.185\linewidth}
          \includegraphics[width=1\linewidth,height=60pt]{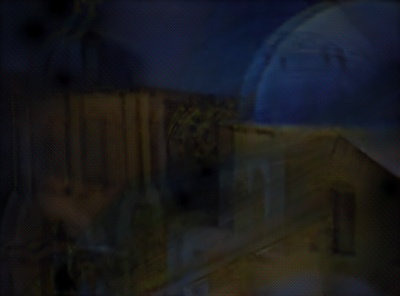}
          \subcaption*{RAGNet \cite{corr/abs-2012-00945}}
     \end{subfigure}
     \begin{subfigure}{0.185\linewidth}
          \includegraphics[width=1\linewidth,height=60pt]{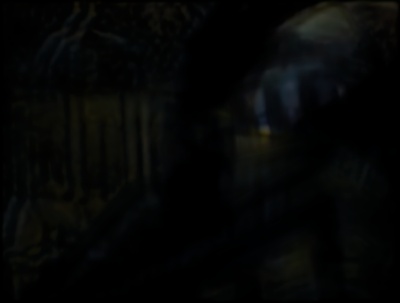}
          \subcaption*{YTMT \cite{nips/HuG21}}
     \end{subfigure}
     \begin{subfigure}{0.185\linewidth}
          \includegraphics[width=1\linewidth,height=60pt]{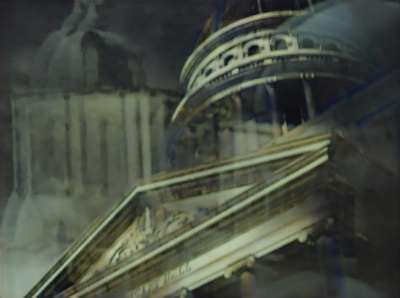}
          \subcaption*{Dong \emph{et al.} \cite{iccv/Dong00BXL21}}
     \end{subfigure}
     \begin{subfigure}{0.185\linewidth}
          \includegraphics[width=1\linewidth,height=60pt]{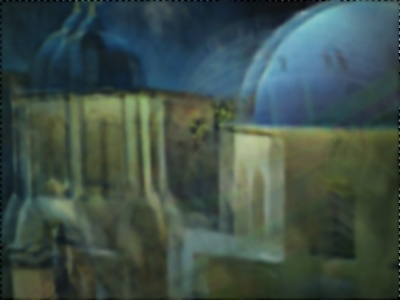}
          \subcaption*{Ours}
     \end{subfigure}
     \begin{subfigure}{0.185\linewidth}
          \includegraphics[width=1\linewidth,height=60pt]{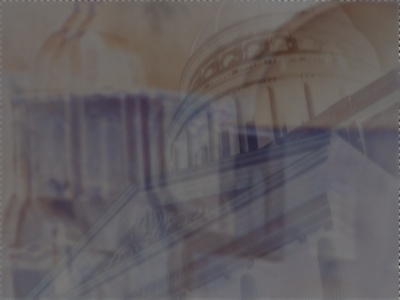}
          \subcaption*{Residue (ours)}
     \end{subfigure}
     \begin{subfigure}{0.185\linewidth}
          \includegraphics[width=1\linewidth,height=60pt]{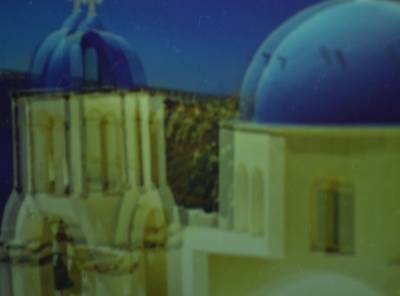}
          \subcaption*{GT}
     \end{subfigure}
      }
     \caption{Visual comparison of estimated reflections between state-of-the-arts and ours on a sample drawn from the $\textrm{SIR}^2$ dataset. Ours has a significant advantage over other alternatives by separating the residue component from the estimations.}
     \label{fig:visual_comp_r}
\end{figure*}

\subsection{Performance Evaluation}

\textbf{Quantitative comparison.}
A comparison is made between state-of-the-art methods as shown in Table \ref{tab:qcomp}, involving Zhang \emph{et al.} \cite{cvpr/ZhangNC18a}, BDN \cite{eccv/YangGLS18}, ERRNet \cite{cvpr/WeiYFW019}, IBCLN \cite{cvpr/LiY0LH20}, RAGNet \cite{corr/abs-2012-00945}, DMGN \cite{tip/FengPJCZL21}, Zheng \emph{et al.} \cite{cvpr/ZhengSCJDK21}, YTMT \cite{nips/HuG21} and ours on four real-world dataset, including Real20 \cite{cvpr/ZhangNC18a} and three subsets of the $\textrm{SIR}^2$ Dataset \cite{iccv/WanSDTK17}.  Besides, to meet the configuration of \cite{iccv/Dong00BXL21}, we train our DSRNet additionally under their training settings, which includes 200 extra real pairs provided by the ``Nature'' dataset \cite{cvpr/LiY0LH20}, and 13,700 synthesized image pairs provided by \cite{cvpr/ZhangNC18a} instead.

It turns out that our method shows its superiority over other competitors on all the testing datasets in both settings, gaining 0.50$dB$ and 1.54$dB$ in terms of average PSNR in the two settings, respectively. Given the four real-world datasets contains a variety of scenes, illumination conditions, and glass thickness, it is not a trivial task to achieve the best performance of all metrics on these datasets simultaneously. The experimental results demonstrate that our proposed SIRS scheme has significantly higher performance and stronger generalization ability, which explains our main contributions.


\textbf{Qualitative comparisons.} To further explain our performance advantages, we provide visual comparisons of transmission layers in Fig.\;\ref{fig:visual_comp} against state-of-the-art methods, including  Zhang \emph{et al.} \cite{cvpr/ZhangNC18a}, BDN \cite{eccv/YangGLS18}, ERRNet \cite{cvpr/WeiYFW019}, IBCLN \cite{cvpr/LiY0LH20}, RAGNet \cite{corr/abs-2012-00945}, YTMT \cite{nips/HuG21}, and Dong \emph{et al.} \cite{iccv/Dong00BXL21}.
As can be observed in Fig.\;\ref{fig:visual_comp}, the method proposed by Zhang \emph{et al.} fails to handle the case containing scattered reflection components in the second row. BDN has trouble dealing with the images with highlights like in the first row and even aggravates the reflections. The problem is likely to be caused by the linear reflection model, which lacks the ability to model specular highlights. For the results of ERRNet, which contains only a single branch to estimate transmission layers, it lacks the direct modeling of the reflections, therefore only removing parts of them. IBCLN also has trouble removing highlights and scattered reflections. The reflection layers in RAGNet are estimated without the participation of the transmission, hence showing inferior performance. YTMT can better cope with the highlights in the last row owing to its dual-stream design but is still limited by the linear assumption and cannot remove strong reflections. The method proposed by Dong \emph{et al.} shows inferior performance in both cases due to the difficulty of estimating the blending weight. Overall, our results are more visually favorable and contain fewer residual reflection components, which further reinforces our claims.

We further deliver a comparison of reflection layers in Fig.\;\ref{fig:visual_comp_r} to illustrate that our method can better reproduce the reflection scenes.  It can be seen that the methods (Zhang \emph{et al.}, RAGNet, and YTMT) based on the additive model provide weak reflection maps. In the meantime, the methods based on the linear model with scalars (BDN and IBCLN) and the one containing an alpha blending map can hardly avoid mixing the transmission components in reflection estimations. This phenomenon reflects the weakness of a fixed simplified model, requiring the reconstruction of more than two components. As shown by our results, introducing a residue term can significantly enhance the restoration quality of the reflection scene, and separates out the components beyond the additive relationship.



\begin{figure*}[t]
     \centering{
     \hspace{-5pt}
     \begin{subfigure}{0.22\linewidth}
          \includegraphics[width=1\linewidth,height=68pt]{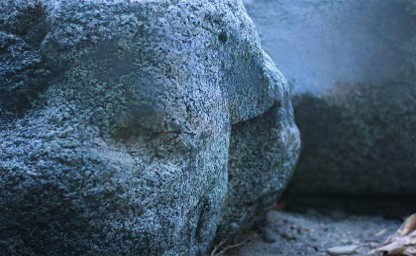}
          \subcaption*{w/o Recons. Loss}
     \end{subfigure}
     \begin{subfigure}{0.22\linewidth}
          \includegraphics[width=1\linewidth,height=68pt]{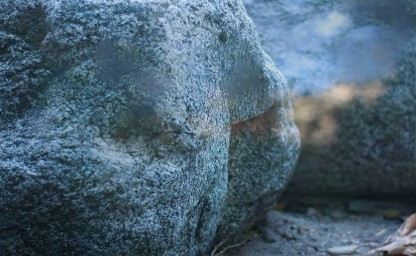}
          \subcaption*{w/ Linear Recons.}
     \end{subfigure}
     \begin{subfigure}{0.22\linewidth}
          \includegraphics[width=1\linewidth,height=68pt]{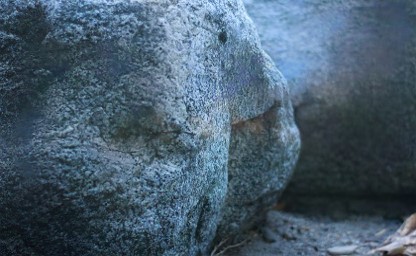}
          \subcaption*{w/o Feature Inter.}
     \end{subfigure}
     \begin{subfigure}{0.22\linewidth}
          \includegraphics[width=1\linewidth,height=68pt]{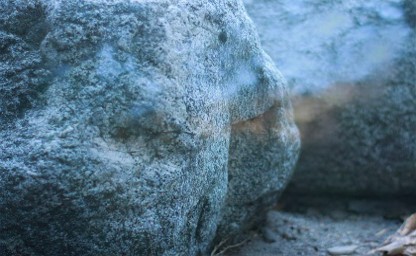}
          \subcaption*{w/ YTMT Inter.}
     \end{subfigure}
     \begin{subfigure}{0.22\linewidth}
          \includegraphics[width=1\linewidth,height=68pt]{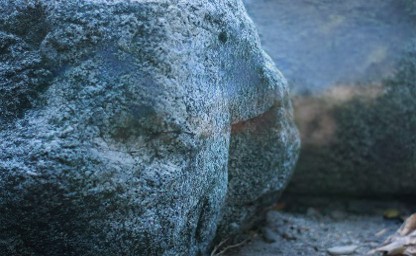}
          \subcaption*{w/o Feature Enc.}
     \end{subfigure}
     \begin{subfigure}{0.22\linewidth}
          \includegraphics[width=1\linewidth,height=68pt]{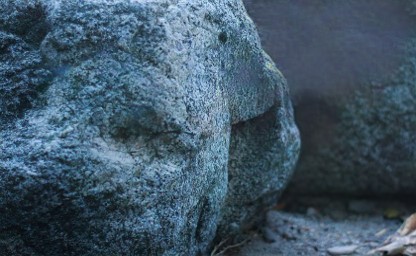}
          \subcaption*{w/ HyperColumn}
     \end{subfigure}
     \begin{subfigure}{0.22\linewidth}
          \includegraphics[width=1\linewidth,height=68pt]{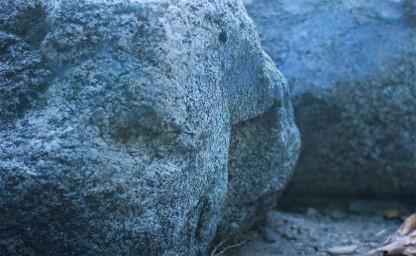}
          \subcaption*{Ours}
     \end{subfigure}
     \begin{subfigure}{0.22\linewidth}
          \includegraphics[width=1\linewidth,height=68pt]{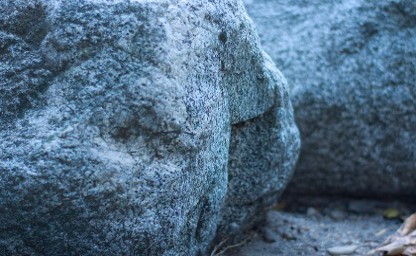}
          \subcaption*{GT}
     \end{subfigure}
      }
     \caption{The visual comparison of different settings involved in the ablation study.}
     \label{fig:ablation}
\end{figure*}


\subsection{Ablation Study}
To better analyze our improvements in the physical model and network structure, we carried out a detailed ablation study from the perspective of the reconstruction losses, feature interaction mechanisms, and feature encoders. We gather the results (PSNR and SSIM averaged on Real20 and $\textrm{SIR}^2$ datasets) in Table \ref{tab:ablation} and deliver a visual comparison in Fig.\;\ref{fig:ablation}.  In what follows we detail every point of them.

\noindent\textbf{Reconstruction Loss.} In view of the problems existing in the linear combination model, we use a learnable residue term to modify it, enhancing its flexibility so that it can cover more reflection scenes. In order to promote the learning of this residue term, we propose the reconstruction loss of residual correction. To illustrate the necessity of this setting, we perform an ablation study to compare the settings without reconstruction loss (w/o Recons. Loss) and using the linear reflection model (w/ Linear Recons.). It shows that the linear model is inferior to the setting without any reconstruction criterion because of the existence of  $\varepsilon$. Meanwhile, without a reconstruction term, the model is likely to have trouble determining the portion of the components, excessively separating content to the reflection layer, resulting in textureless regions in the transmission layer as shown in Fig.\;\ref{fig:ablation}.

\begin{table}[t]
    \centering
    \begin{tabular}{lcc}
    \toprule[1pt]
    \multicolumn{1}{c}{Models}  & PSNR & SSIM \\ \hline
    w/o Recons. Loss & 24.84 & 0.897 \\
    w/ Linear Recons. & 22.21 & 0.881 \\ 
    w/o Feature Inter.  & 24.81 & 0.895 \\
    w/ YTMT Inter. & 24.89 & 0.901 \\ 
    w/o Feature Enc. & 24.75 & 0.903 \\
    w/ HyperColumn  & 23.99 & 0.894 \\ 
    \multicolumn{1}{c}{Ours} & \textbf{25.40} & \textbf{0.905} \\
    \bottomrule[1pt]
    \end{tabular} 
\caption{Ablation study on different configurations. The initial two rows compare the configuration devoid of any reconstruction loss and that integrating a linear reconstruction loss. Rows 3-4 offer a comparison between the settings lacking feature interaction and incorporating the YTMT feature interaction mechanism. Rows 5-6 shed light on configurations that exclude the feature encoder and utilize HyperColumn as the feature encoder. The last row shows the results of  our full version. }
\label{tab:ablation}
\vspace{-10pt}
\end{table}

\noindent\textbf{Feature Interaction Mechanism.} The feature interaction mechanism connects the two information streams in our proposed DSRNet, both exchanging and conditioning the dual-stream information, which  facilitates the layer reconstruction. As shown in Table\;\ref{tab:ablation}, the model performance appears to degrade after removing the feature interaction (w/o Feature Inter.), which tells us that the network can hardly achieve state-of-the-art performance via a two-branch network without feature interaction. Further, we include the results of the YTMT feature interaction mechanism (w/ YTMT Inter.) by replacing MuGI with the YTMT mechanism. It can be seen that the model with the YTMT mechanism is only slightly better than the scheme without interaction, which demonstrates the superiority of the proposed MuGI mechanism. 


\noindent\textbf{Feature Encoder.} In comparison to the hypercolumn framework introduced by the previous method (referred to as "w/ HyperColumn") and the approach that without using any feature encoder ("w/o Feature Enc."), our method demonstrates noteworthy performance advantages. These advantages stem from two main factors. Firstly, our approach employs a hierarchical and gradual reduction of channel dimensions, followed by fusion operations. Secondly, DSFNet leverages a feature interaction mechanism, allowing high-level semantic features to interact before being fed into the subsequent stage. 

\section{Conclusion}
In this paper, we modified the commonly-used linear model in the single image reflection separation task and proposed the reconstruction loss with residual correction. The proposed model is more flexible and effective compared with the previous methods. Meanwhile, we further improved the feature interaction mechanism in dual-stream networks and the usage of hierarchical semantic information, proposed the MuGI as a novel interaction paradigm, and a dual-stream semantic-aware network, namely DSRNet. Extensive experiments revealed that our proposed method has achieved state-of-the-art performance on all real-world benchmark datasets and verified our contributions. In the future, further constraints are desired to be applied to the residue term to reduce its solution space, and more reflection models are hopefully to be covered by it to fit a wider range of reflection phenomena in the real world.


{\small
\bibliographystyle{ieee_fullname}
\bibliography{egbib}
}

\end{document}